\documentclass[10pt,twocolumn,letterpaper]{article}
 \pdfoutput=1
\usepackage{cvpr}
\usepackage{times}
\usepackage{epsfig}
\usepackage{graphicx}
\usepackage{amsmath}
\usepackage{amssymb}
\usepackage{bm}
\usepackage{subcaption}
\usepackage{slashbox}
\usepackage{color}
\usepackage[english]{babel}
\usepackage[linesnumbered,ruled,vlined]{algorithm2e}
\usepackage[breaklinks=true,bookmarks=false]{hyperref}

\cvprfinalcopy

\def\cvprPaperID{16} 

\begin{document}
\title{Kernalised Multi-resolution Convnet for Visual Tracking}

\author{Di Wu, Wenbin Zou\thanks{Corresponding author.}, Xia Li\\
Shenzhen University\thanks{Shenzhen Key Lab of Advanced Telecommunication and Information Processing, College of Information Engineering, Shenzhen University.}\\
{\tt\small dwu, wzou, lixia@szu.edu.cn}
\and
Yong Zhao\\
Peking University Shenzhen Graduate School\\
{\tt\small yongzhao@pkusz.edu.cn}
}

\maketitle
\begin{abstract}
    Visual tracking is intrinsically a temporal problem.
    Discriminative Correlation Filters (DCF) have demonstrated excellent performance for high-speed generic visual object tracking.
    Built upon their seminal work, there has been a plethora of recent improvements relying on convolutional neural network (CNN) pretrained on ImageNet as a feature extractor for visual tracking. However, most of their works relying on ad hoc analysis to design the weights for different layers either using boosting or hedging techniques as an ensemble tracker.
    In this paper, we go beyond the conventional DCF framework and propose a Kernalised Multi-resolution Convnet (KMC) formulation that utilises hierarchical response maps to directly output the target movement.
    When directly deployed the learnt network to predict the unseen challenging UAV tracking dataset without any weight adjustment, the proposed model consistently achieves excellent tracking performance.
    Moreover, the transfered multi-reslution CNN renders it possible to be integrated into the RNN temporal learning framework, therefore opening the door on the end-to-end temporal deep learning (TDL) for visual tracking.

\end{abstract}
\section{Introduction}
\begin{figure}[t]
        \centering
        \includegraphics[width=0.45\textwidth]{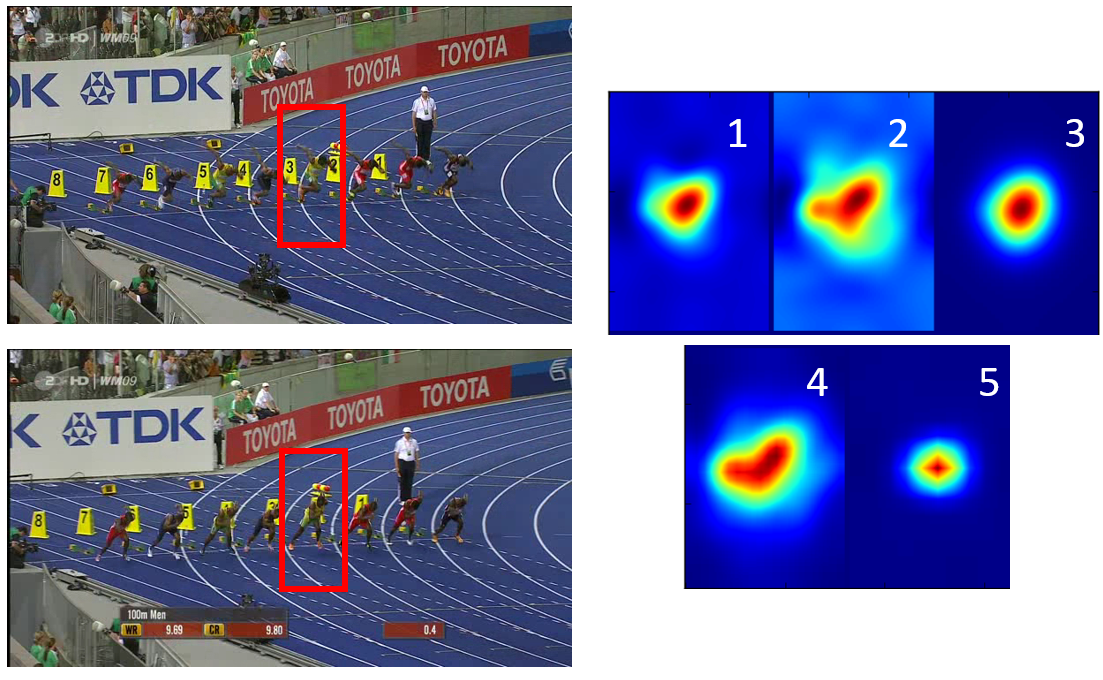}
        \caption{ {\small Response maps generated from hierarchical deep features. Left: two frames from the ``Bolt" sequence, red boxes are image patches with padding added. Right: hierarchical response maps from correlation filters (response maps have been resized and rescaled to the same shape and scale). It can be seen that because different layers of deep features encode different semantics, the maximum response points from different layers also vary.}}
        \label{fig:response_map}
\end{figure}

Visual tracking is the task of predicting the trajectory of a target in a video. The task of tracking, a crucial component of many computer vision systems, can be naturally specified as an online learning problem.
This paper focuses on the challenging problem of monocular, generic, realistic object tracking.

Visual object tracking has recently witnessed substantial progress due to powerful features extracted using deep convolutional neural networks.
For online applications, one simple approach to transfer offline pre-trained CNN features is to add one or more randomly initialised CNN layers, named as adaptation layers, on top of the pre-trained CNN model. However, as~\cite{wang2016stct}  empirically observe that this transfer learning method suffers from severe over-fitting because of the limited training data.
It is also observed in ~\cite{wu2016deep, wu2014leveraging} that the differences in mean activations in intermediate layers have noticeable effect for multi-model fusion.
The online learnt parameters mainly focus on recent training samples and are less likely to be well generalised to both historical and future samples. This phenomenon can be fatal to online visual tracking where the target often undergoes significant appearance changes or heavy occlusion.
CNNs also have shown impressive performance as an offline feature extractor for tracking~\cite{NIPS2013_5192, ma2015hierarchical} in lieu of traditional handcrafted features (\emph{e.g.}, HOG, Color Moment). Features from these deep convolutional layers are discriminative while preserving spatial and structural information.

The marriage between DCF~\cite{henriques2012exploiting}, which has the advantage of being efficient in training translational images in the fourier space, and deep features, which excel at image representation, further advances the visual tracking community~\cite{qi2016hedged, danelljan2016eccv}.
However, there are some conflicting conclusions concerning the representation power from different layers of CNN: shallow convolutional layers have been highlighted in the DCF-based methods~\cite{danelljan2015iccvworkshop} whereas~\cite{qi2016hedged} found that the performance generally increases as layer depth is increased hence only convolutional layers deeper than 10th is used. Fig.~\ref{fig:response_map} shows the response maps generated from hierarchical deep features. Since different layers from convnet capture different semantic meanings, maximum translational locations obtained via different convnet layers could vary: low level layers excel at discriminating intra-class variations whereas high level layers excel at discriminating inter-class variations. Hence, for generic object tracking, it is imperative to combine features from different layers to have a generalised representation from hierarchies of response maps.

Visual tracking is intrinsically also a temporal problem, but many previous approaches~\cite{bolme2010visual, wang2016stct} mainly focus on the design of a robust appearance model. An end-to-end object tracking approach which directly maps from raw sensor input to object tracks in sensor space is proposed in~\cite{Ondruska:2016:DTS:3016100.3016374}. Their proposed method works well in a simulated environment, but the applicability on the realistic RGB imagery dataset is worth further investigation.
On reason that temporal deep learning based techniques are challenging to be employed in realistic settings is due to the difficulty for recurrent net to have robust, meaningful temporal input.

In this paper, we propose a kernelised multi-resolution convnet tracking algorithm that utilises the intermediate response maps from the correlation filter output and learns the implicit translational output from the multi-resolution convnet.
The key contributions can be summarised as follows:
\begin{itemize}
\item We incorporate kernalised form into CNN feature representations for the non-linear regression tasks that consistently outperforms its linear counterpart, concluding the kernalised version should be the preferred choice in the process of designing correlation filters.
\item We learn a novel representation for multi-resolution response maps generated from different layers of a pre-trained CNN, negating the need to design the weights for various convnets layers. Moreover, the learnt representation renders it possible to be included into the end-to-end temporal deep learning (TDL) pipeline.
\item We use an adaptive hedge method to update the model learning rate, taking the model stability into consideration, making the update of target model smoother and more robust.
\end{itemize}

\section{Related Work}
\begin{figure*}[t]
        \centering
        \includegraphics[width=0.8\textwidth]{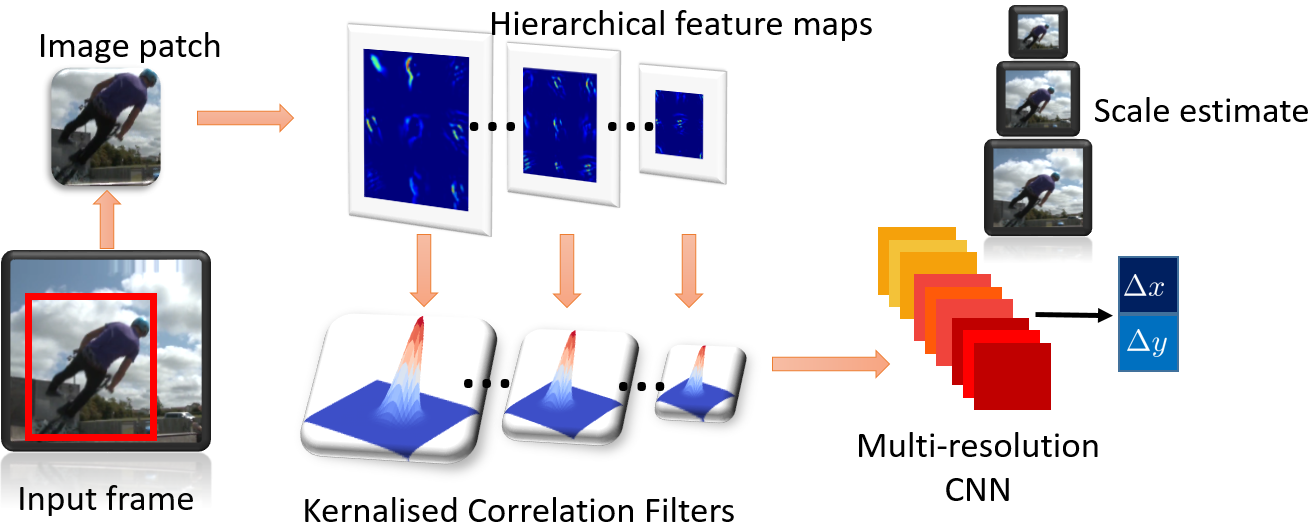}
        \caption{{\small The proposed algorithm overview: we first extract hierarchical CNN features from the image patch of interest; then we project deep features into kernel space for correlation filters; the target translational movement is predicted via multi-resolution convnets and later we update the scale and appearance models using an adaptive learning rate scheme.}}
        \label{fig:overview}
\end{figure*}

We review recent tracking methods closely related to this work as in the following three sections.

\vspace{\baselineskip}
\noindent \textbf{Correlation filter based trackers:} Most recent works~\cite{bolme2010visual, wang2016stct} strives for a better appearance models for the tracker. DCF~\cite{bolme2010visual, henriques2012exploiting} have demonstrated excellent performance for high-speed visual object tracking. The key to their success is by observing that the resulting data matrix of translated patches is circulant, the cyclic shifts could be diagonalised with the Discrete Fourier Transform, reducing both storage and computation to obtain the next frame response map. However, most of these trackers depend on the spatiotemporal consistency of visual cues. Therefore, they can handle mostly short-term tracking or object with static appearance.

\vspace{\baselineskip}
\noindent \textbf{Deep feature based trackers:} Recent works exploit the structure of CNN to learn the target online: a three-layer CNN is trained on-the-flight in~\cite{li2014robust}; a deep autoencoder~\cite{wang2013learning} is first pre-trained offline and then finetuned for binary classification in online tracking. Since the pre-training is performed in an unsupervised way by reconstructing gray images with very low resolution, the learned deep features has limited discriminative power for tracking. Moreover, without pre-traning and with limited training samples obtained online, CNN fails to capture object semantics and is not robust to deformation.
Both~\cite{li2014robust} and~\cite{wang2013learning} train deep networks online with limited training samples, and inevitably suffer from overfitting.
 Transferring the hierarchical features learned for image classification tasks have been shown to be effective for numerous vision tasks, e.g., image segmentation~\cite{long_shelhamer_fcn}, salient object detection~\cite{zou2015harf}.
 More recent methods~\cite{wang2016stct, wang2015visual, hong2015online, ma2015hierarchical} adopt deep convolution networks trained on a large scale image classification task~\cite{ILSVRC15} to improve tracking performance.  The rich representation of transferred features from deep nets enables trackers to construct more robust, power appearance model over the traditional hand crafted feature based trackers.

\vspace{\baselineskip}
\noindent \textbf{Spatio-temporal model based trackers:} Variations in the appearance of the object in tracking, such as variations in geometry/photometry, camera viewpoint, partial occlusion or out-of-view, pose a major challenge to object tracking.  TLD~\cite{kalal2012tracking} employs two experts to identify the false negatives and false positives to train a detector. The experts are independent, which ensures mutual compensation of their errors to alleviate the problem of drifting. A short and long term cognitive psychology principle is adopted in~\cite{hong2015multi}  to design a flexible representation that can adopt to changes in object appearance during tracking. A parameter-free Hedging algorithm is proposed in~\cite{chaudhuri2009parameter} for the problem of decision-theoretic online learning, especially for the applications when the number of actions is very large and optimally setting the parameter is not well understood. An improved Hedge algorithm considering historical performance is proposed in~\cite{qi2016hedged} to weight the decision from different CNN layers.

\section{Proposed Algorithm}

As shown in Fig.~\ref{fig:overview}, the proposed approach consists of three steps: extracting hierarchical CNN features, projecting features into kernel space for correlation filters, predicting target translational movement via multi-resolution convnet.
We first review the correlation filter as our building block. Then we present the technical details of the projection of deep feature to kernel space, the model of learning translational output using a multi-resolution convnets and an adaptive learning rate scheme for model update.

\vspace{\baselineskip}
\noindent \textbf{Correlation Filter:} Correlation filters based trackers~\cite{bolme2010visual, henriques2012exploiting, danelljan2014accurate, henriques2015high} exploit the circulant structure of training and testing samples to greatly accelerate the training and testing process.
Let $\textbf{X}^k \in \Re^{ D \times M \times N}$ denotes the feature sets where $D$ denotes the number of feature maps; $M, N$ denote the shape of feature maps and $k$ denotes the $k$-th input feature map extracted from the $k$-th convolutional layers from a pretrained CNN; $\textbf{Y} \in \Re^{M \times N}$ denotes a gaussian shaped label matrix which is subject to a 2D Gaussian distribution with zero mean and standard deviation proportional to the target size.
The goal of training is to find a set of filters $\textbf{W}^k$ that minimises the squared error over sets of circulant translated samples $\textbf{X}^k$ and their regression targets $\textbf{Y}$:
\begin{equation}
 \textbf{W}^k = \underset{\textbf{W}^k}{\mathrm{argmin}} \lVert \textbf{Y} - \textbf{X}^k \bullet \textbf{W}^k \rVert^2 + \lambda \lVert \textbf{W}^k \rVert^2
\end{equation}
where
\begin{equation}
\textbf{X}^k \bullet \textbf{W}^k = \sum^D_{d=1} \textbf{X}^k_d \odot \textbf{W}^k_d
\end{equation}
with the symbol $\odot$ denotes element-wise product.
\noindent The minimizer has a closed-form:
\begin{equation}
 \textbf{W} = \frac{\textbf{X}^T\textbf{Y}}{\textbf{X}^T\textbf{X} + \lambda \textbf{I}}
 \end{equation}
where $\textbf{I}$ is an identity matrix and $\lambda$  is a regularisation parameter that controls overfitting. We drop the superscript $k$ for notational simplicity.
In general, a large system of linear equations must be solved to compute the solution, which can become prohibitive in a real-time setting.
With training data being cyclic shifts patches, all operations can be done element-wise on their diagonal elements~\cite{henriques2015high}.

The filter can be modelled in the Fourier domain by:
\begin{equation}
 \bm{\mathcal{W}} = \underset{\bm{\mathcal{W}}}{\mathrm{argmin}} \lVert \bm{\mathcal{Y}} - \bm{\mathcal{X}} \bullet \bm{\mathcal{W}}^k \rVert^2_F + \lambda \lVert \bm{\mathcal{W}} \rVert^2_F
  \label{eq:linearw}
\end{equation}
where
\begin{equation}
\bm{\mathcal{X}} \bullet \bm{\mathcal{W}} = \sum^D_{d=1} \bm{\mathcal{X}}_d \odot \bm{\mathcal{W}}_d
\label{eq:feature_additon}
\end{equation}

The corresponding minimizer in the Fourier domain has the closed-form:
\begin{equation}
 \bm{\mathcal{W}} = \frac{\bm{\mathcal{X}}^{\ast} \odot \bm{\mathcal{Y}}}{(\bm{\mathcal{X}}^{\ast} \odot \bm{\mathcal{X}} + \lambda \bm{\mathcal{I}})}
\end{equation}
where $\ast$ denotes the Hermitian transpose and since diagonal matrices are symmetric, taking the Hermitian transpose only left behind a complex-conjugate.

\subsection{Kernelisation of Deep transfered features}
\label{sec:kernelisation}

In Kernelized Correlation Filter (KCF)~\cite{henriques2015high}, the feature $\textbf{X}$ is mapped to a Hilbert space $\phi (\textbf{X})$.
By employing a kernel $\kappa(\textbf{X}, \textbf{X}') = \langle \phi (\textbf{X}), \phi (\textbf{X}') \rangle$, Eq.~\ref{eq:linearw} becomes:
\begin{equation}
\bm{\mathcal{W}} = \underset{\bm{\mathcal{W}}}{\mathrm{argmin}} \lVert \bm{\mathcal{Y}} - \langle \phi (\bm{\mathcal{X}}) , \bm{\mathcal{W}} \rangle \rVert^2_F + \lambda \lVert \bm{\mathcal{W}} \rVert^2_F
 \label{eq:kernelw}
\end{equation}

The power of the kernel trick comes from the implicit use of a high-dimensional feature space $\phi(\textbf{X})$ without ever instantiating a vector in the space. Even though the regression function's complexity grows with the number of samples which is the major drawback of the kernel trick, assuming circulant data and the adopted kernel being shift invariant, the kernel correlation can be computed efficiently.
In our experiment, an radial basis function (RBF) kernel, which satisfies the shift invariant property, is adopted:
\begin{equation}
\bm{k}^{\bm{XX}'} = exp(-\frac{1}{\sigma}(\lVert \bm{X} \rVert ^2 + \lVert \bm{X}' \rVert ^2 - 2 \mathcal{F}^{-1} (\bm{\mathcal{X}}^{\ast} \odot \bm{\mathcal{X'}} ))
 \label{eq:kernel_w_2}
\end{equation}
where $\mathcal{F}^{-1}$ denotes the Inverse DFT and the full kernel correlation can be computed in only $\mathcal{O}(n \log n)$ time.

Expressing the solution $\bm{\mathcal{W}}$ as a linear combination of the samples: $\bm{\mathcal{W}} = \sum_i \alpha_i \phi(\bm{\mathcal{X}}_i)$ renders an alternative representation $\bm{\alpha}$ to be in the dual space, as opposed to the primal space $\bm{\mathcal{W}}$ (Representer Theorem~\cite{scholkopf2001learning}). The variables under optimisation are thus  $\bm{\alpha}$:
\begin{equation}
\bm{\alpha} = (\bm{k}^{\bm{XX}'} + \lambda \bm{\mathcal{I}}) ^{-1} \bm{\mathcal{Y}}
 \label{eq:alpha}
\end{equation}

For detection stage for translation estimation, given an image patch feature $\textbf{Z}^k \in \Re^{ D \times M \times N}$, the response map  is obtained by:
\begin{equation}
R(\bm{Z}^k) = \mathcal{F}^{-1} (\mathcal{F}(\bm{k}^{\bm{XZ}^{k'}}) \odot \mathcal{F} (\bm{\alpha}^k))
 \label{eq:response_map}
\end{equation}

The model for frame $t$ is updated with learning rate $\eta$ as:
\begin{equation}
\bm{\alpha_{t}^k} = (1- \eta) \bm{\alpha_{t-1}^k} + \eta \bm{\alpha}^k
 \label{eq:model_update}
\end{equation}

\begin{figure}[t]
        \centering
        \includegraphics[width=0.5\textwidth]{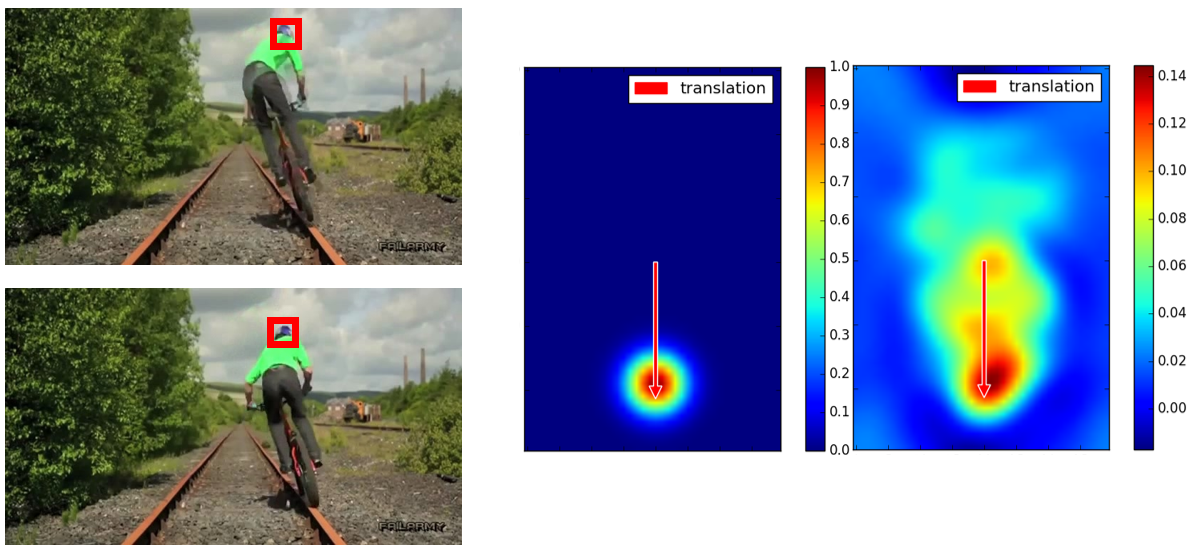}
        \caption{ {\small Response maps for one frame of the ``Biker" sequence. Left: two consecutive frames; middle: ideal response map (2D gaussian); right: the actual response map. Arrows represent the translational vector. Due to various factors, \emph{e.g.}, motion blur, target appearance change, scale change, boundary effect, the actual response map is not a univariate gaussian and the maximum response does not strictly correspond to the translational movement.}}
        \label{fig:response_map_biker}
\end{figure}

\subsection{Decoding Response map using Multi-Res CNN}
\label{sec:multi_res_cnn}

Previous works estimate the next frame location via the maximum response point on the response map. We argue that albeit the maximum response point carries physical meaning of the translational information between two consecutive frames, the precondition is that of static target appearance, \emph{i.e.}, the maximum response map corresponds to the cyclic shifts of appearance unchanged target. Due to subtle changes of target appearance, \emph{e.g.}, rotation, scale, cyclic boundary effect or motion blur of two consecutive frames, the accuracy of finding the translation using maximum response could be compromised (\emph{c.f.} Fig.~\ref{fig:response_map_biker}).

Moreover, when using deep features from the pre-trained convnets for the input representation $\textbf{X}$, the hierarchies of feature maps capture different semantic information(\emph{c.f.} Fig.~\ref{fig:response_map}):  features from deep layers capture rich category level semantic information, which is useful for object classification, but they are not the optimal representation for visual tracking because spatial details captured by earlier layers are also important for accurately localising the targets. On the other hand, as the features in the earlier layers capture low level visual characteristics and are more class-generic rather than discriminative as ones in the later layers, methods based on features from earlier layers are likely to fail in challenging scenarios when the target size is small and high level object is the target of tracking.

\begin{figure}[t]
        \centering
        \includegraphics[width=0.35\textwidth]{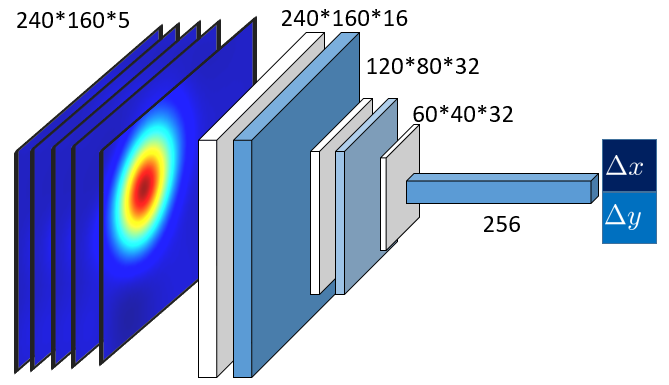}
        \caption{{\small A multi-resolution convnet is deployed to decode the hierarchical response maps.}}
        \label{fig:multi_cnn}
\end{figure}

We interpret the response maps from the hierarchies of convolutional layers as a nonlinear counter part of an image pyramid representation and employ a learning framework for mapping the hierarchical outputs to target translation.
A multi-resolution neural network (\emph{c.f.} Fig.~\ref{fig:multi_cnn}) is deployed to exploit the implicit translational information from hierarchies response maps:

\begin{equation}
\Delta x, \Delta y \leftarrow \emph{convnets} (R(\bm{Z}^k))
\label{eq:response_map-convnet}
\end{equation}

The loss of the convnets is set to be the root mean square (RMS) of the normalised translational movement $(\Delta x, \Delta y)$ and the predicted movement $(\Delta x', \Delta y')$:
\begin{equation}
L_{pos} = \sqrt{ \frac{1}{2} (\parallel  \Delta x - \Delta x' \parallel ^2_2 + \parallel  \Delta y - \Delta y' \parallel ^2_2)}
\label{eq:loss}
\end{equation}
The proposed formulation enables efficient integration of multi-resolution response maps to decode translational information implicitly and demonstrates its competitiveness over the corresponding counterpart baselines that using the mean of maximum responses as the indicator for target movement.

\subsection{Adaptive learning rate}
\label{sec:adl}
Traditional correlation based filter updates the model by a fixed parameter $\eta$.  For generic object tracking, however, there are two crucial factors for model update using deep features:
(1) the appearance of object of interest usually changes at irregular pace (sometimes gently and sometimes vehemently). This means that the scale of the learning rate should reflect the appearance change of the target;
(2) depending on whether the object of interest being a low level visual cues or high level object entity, the hierarchies of response maps render correspondingly different maximum response values.

Let the ultimate target position is predicted $(x_p, y_p)$ at time $t$, each layer $k$ will incur a loss from its response map $R^k_t$:
\begin{equation}
l^k_t = max(R^k_t) - R^k_t(x_p, y_p)
\label{eq:adl_loss}
\end{equation}

We then measure the stability of layer $k$ at time $t$:
\begin{equation}
s^k_t = \frac{| l^k_t - \mu^k_t  |}{\sigma^k_t}
\label{eq:adl_stability}
\end{equation}
where $ \mu^k_t, \sigma^k_t$ are the mean and variance for the loss $l^k_t$ during time period $\Delta t$.

A larger $s^k_t$ indicates the layer is less correlated with the object at frame $t$ hence the model update $\eta$ should be decreased as well. Therefore, we propose an adaptive learning rate for different layers of model update that is linear to the model stability as:
\begin{equation}
    \eta^k = s^k_t \times \eta
\label{eq:adl_stability}
\end{equation}

Algorithm~\ref{Algorithm:KMC} summarises the main steps of the proposed approach for visual tracking.

\begin{algorithm}[t]
\DontPrintSemicolon 
\KwIn{target position $(x_1, y_1)$ and size $(w_1, h_1)$, first tracking frame;}
\KwOut{predicted target positions $(x_t, y_t)$ and sizes $(w_t, h_t)$ in the following frames.}

\For{t = 1, 2, \ldots} {
  Crop target images with padding; \;
  Extract deep features from layers before max pooling layers in VGG-Net19 to obtain 5 hierarchical representations $\textbf{X}$; \;
  Project deep features into kernel space~\ref{sec:kernelisation}; \;
    \uIf{$t = 1$}{
      Obtain deep kernel model $\bm{\alpha}$~ (Eq.\ref{eq:alpha}); \;
    }
    \Else{
      Update target position prediction by the trained multi-resolution convnet~\ref{sec:multi_res_cnn}; \;
      Update model scale using DSST~\cite{danelljan2017dsst_tpami}; \;
      Update kernel model using adaptive learning rate~\ref{sec:adl}; \;
    }
}
\caption{Kernalised Multi-resolution Convnet}
\label{Algorithm:KMC}
\end{algorithm}
\section{Experiments}
\begin{table*}[t]
\small
   \centering
        \begin{tabular}{|l|| *{2}{c}| *{2}{c} |}\hline
            \backslashbox{{\small Method}}{{\small Dataset}} & \multicolumn{2}{|c|}{\small OTB-2013}  & \multicolumn{2}{|c|}{\small OTB-2015}  \\ \hline \hline
                                                    & P20       & AUC       & P20       & AUC  \\ \hline \hline

            {\small DeepDCF }                       &  58.0     &  38.4     & 64.3      & 44.0   \\\hline
            {\small \textbf{DeepKCF} }  &  66.2 ${\color{red}(8.2\uparrow)}$    &  46.1 ${\color{red}(7.7\uparrow)}$    &  71.4 ${\color{red}(7.1\uparrow)}$     &  49.1 ${\color{red}(5.1\uparrow)}$\\\hline \hline
            {\small MaxRes }                        &  57.5     &  38.8     &  65.8     &  45.9  \\\hline
            {\small \textbf{Multi-Resolution CNN} }        &  70.6${\color{red}(13.1\uparrow)}$     &  46.7  ${\color{red}(7.9\uparrow)}$   &  77.8 ${\color{red}(12.0\uparrow)}$    &  53.6 ${\color{red}(7.7\uparrow)}$ \\\hline\hline
            {\small Fixed Learning Rate}           &  74.0${\color{red}(3.4\uparrow)}$      &  51.8${\color{red}(5.1\uparrow)}$      &  79.8${\color{red}(2.0\uparrow)}$      &  57.0${\color{red}(3.4\uparrow)}$    \\\hline
            {\small \textbf{Adaptive Learning Rate}}&  74.2 ${\color{red}(0.2\uparrow)}$    &  52.1${\color{red}(0.3\uparrow)}$     &  79.9${\color{red}(0.1\uparrow)}$     &  57.4${\color{red}(0.4\uparrow)}$\\\hline

        \end{tabular}

    \caption{ {\small
          Baseline comparisons with arrows indicating performance changes comparing with the upper row parallel experiment.
          First two rows investigate the impact of projecting deep features into a kernel space \emph{vs.} using linear deep features.
          Next two rows compare the traditional using max location \emph{vs.} the proposed multi-resolution convnets paradigm for decoding target translational movement.
          Last two rows illustrate the benefit of adaptive learning rate through time (with scale update). }
          } \label{table_baseline}
\end{table*}

We validate our proposed KMC framework by evaluating two genres of experiments: one compares with correlation filter based baseline trackers and one compares with several state-of-the-art trackers. We perform comprehensive experiments on three datasets: OTB-2013~\cite{wu2013online}, OTB-2015~\cite{wu2015object} and UAV123~\cite{mueller2016benchmark}.

\vspace{\baselineskip}
\noindent \textbf{Implementation details:}
For feature extraction, we crop the image patch with $2.2$ padding size and resize the image patch to $240*160$ because the average target size is of ratio $\frac{3}{2}$.
Feature bandwidth $\sigma$ in Eq.~\ref{eq:kernel_w_2} is $0.2$.
The learning rate $\eta$ for Eq.~\ref{eq:adl_stability} is $0.0025$ which is only one fifth of ones chosen in~\cite{qi2016hedged} because of the model is in kernel space and tends to be more stable.
After the forward propagation, we use the VGG-Net with 19 layers and five outputs before the max pooling layers (\emph{i.e.}, $`block1\_conv2'$, $`block2\_conv2'$, $`block3\_conv4'$, $`block5\_conv4'$). Note that instead of heuristically choosing layers as in~\cite{qi2016hedged, danelljan2016eccv}, we have included all layer outputs before the max pooling layers. This chosen methodology makes the approach more generic and less dataset dependant.

The convnet in Sec.~\ref{sec:multi_res_cnn} is trained on OTB-2015 dataset. Note that albeit trained on the OTB-2015 dataset, the convnet during detection will not be able to see the exactly same response maps twice unless exact location update is achieved. Hence it requires the convnet to be able to generalise to unseen hierarchical response maps and predict reasonable translational output. Furthermore, we verify the generalisation of the KMC over HDT~\cite{qi2016hedged} tracker on the unseen UAV123 dataset.

For scale estimation, we use a scale pyramid representation as in~\cite{danelljan2017dsst_tpami} with 11 scales and a relative scale factor $1.02$. The updates consistently helps more accurate update of target models while maintaining the computational cost low.

\vspace{\baselineskip}
\noindent \textbf{Evaluation Methodology:}
Following the evaluation strategy  of ~\cite{wu2013online}, all trackers are compared using two measures: precision and success. Precision is measured as the distance between the centres of the ground truth bounding box and the corresponding tracker generated bounding box. The precision plot shows the percentage of tracker bounding boxed within a given threshold distance in pixels of the ground truth. To rank the trackers, the conventional threshold of 20 pixels (\emph{P20}) is adopted. Success is measured as the intersection over union of pixels. The success plot shows the percentage of tracker bounding boxes whose overlap score is larger than a given threshold and the trackers are ranked according to the Area Under Curve (\emph{AUC}) criteria. All sequences are evaluated using One-Pass Evaluation (OPE) as in~\cite{wu2013online}.

\subsection{Baseline comparison}

\noindent \textbf{Benefit of Kernalising deep features:}
We first evaluate the impact of projecting deep features into a kernel space (\emph{DeepKCF}) vs. using the traditional linear projection (\emph{DeepDCF}) in first two rows of Tab.~\ref{table_baseline}. The deep features used are the layer from $`block3\_conv4'$. It can be seen that projecting deep feature into kernel space consistently yields a noticeable increase in both performance measures. The computation cost for kernel projection is $\mathcal{O}(n \log n)$  and for linear kernel is $\mathcal{O}(n)$. Therefore, given marginal increase in computational cost, we conclude that for constructing appearance model, kernel projection is the preferred methodology versus the traditional linear projection.

\vspace{\baselineskip}
\noindent \textbf{Benefit of multi-resolution CNN:}
We then evaluate the impact of decoding response map using a multi-resolution CNN (\emph{Multi-Resolution CNN}) vs. directly using the max location information (\emph{MaxRes}) for decoding target location from hierarchical response maps in Tab.~\ref{table_baseline}. \emph{MaxRes} use the maximum position from the mean of hierarchical response maps with later response maps being resized to the largest first layer response map of size $240*160$. The performance of \emph{MaxRes} is even worse than using single layer, signifying the need to intelligently combine multiply layer output. The learnt convnets consistently performs better than that of \emph{MaxRes} for a large margin. Moreover, the learning paradigm opens the door for the integration of recurrent network into the pipeline.

\vspace{\baselineskip}
\noindent \textbf{Benefit of adaptive learning rate through time:}
We also investigate the scheme of adjusting the learning rate adaptively vs. fixing the learning rate through time in last two rows of Tab.~\ref{table_baseline}. Consistent improvements across two metrics are observed. The margin of improvement, however, is less prominent. We reason that in part, it is due to almost saturate performance; on the other hand, an end-to-end learning scheme for model update could be more preferable and should be further investigated.

\subsection{Comparisons with the-state-of-the-art trackers}
\setlength{\fboxsep}{1pt}%
\setlength{\fboxrule}{1pt}%
\begin{figure*}[t]
       \centering
        \begin{subfigure}[c]{0.4\textwidth}
        \centering
        \includegraphics[width=7cm,height=5cm, clip]{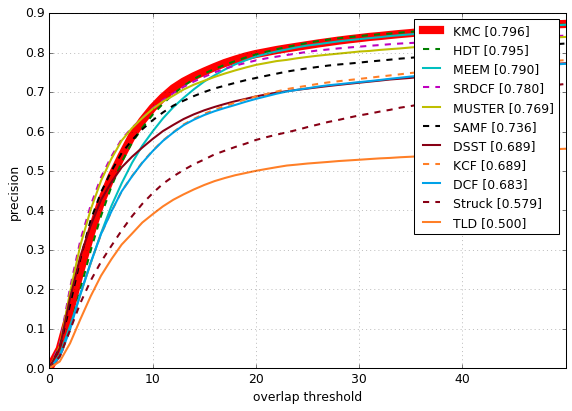}
        \caption{\small{Precision plot for OTB-2015 Dataset}}
        \end{subfigure}%
        \begin{subfigure}[c]{0.4\textwidth}
        \centering
            \includegraphics[width=7cm,height=5cm, clip]{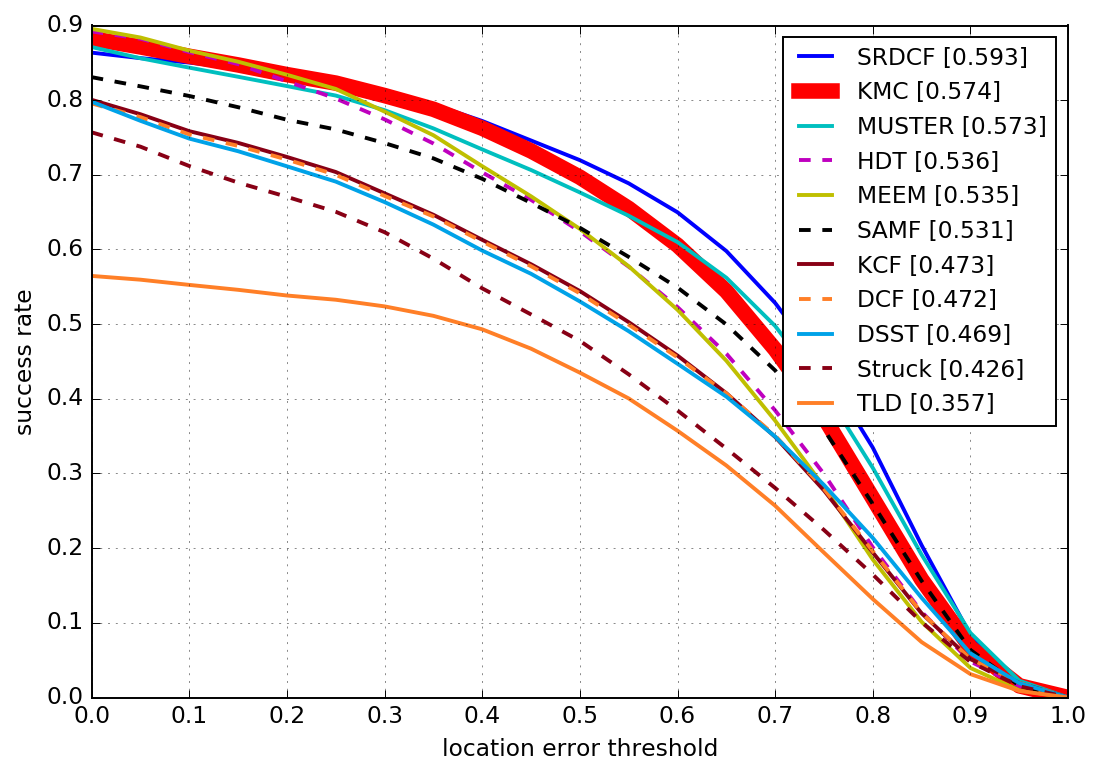}
            \caption{\small{Success plot for OTB-2015 Dataset}}
        \end{subfigure}
        \\
         \begin{subfigure}[c]{0.4\textwidth}
        \centering
        \includegraphics[width=7cm,height=5cm, clip]{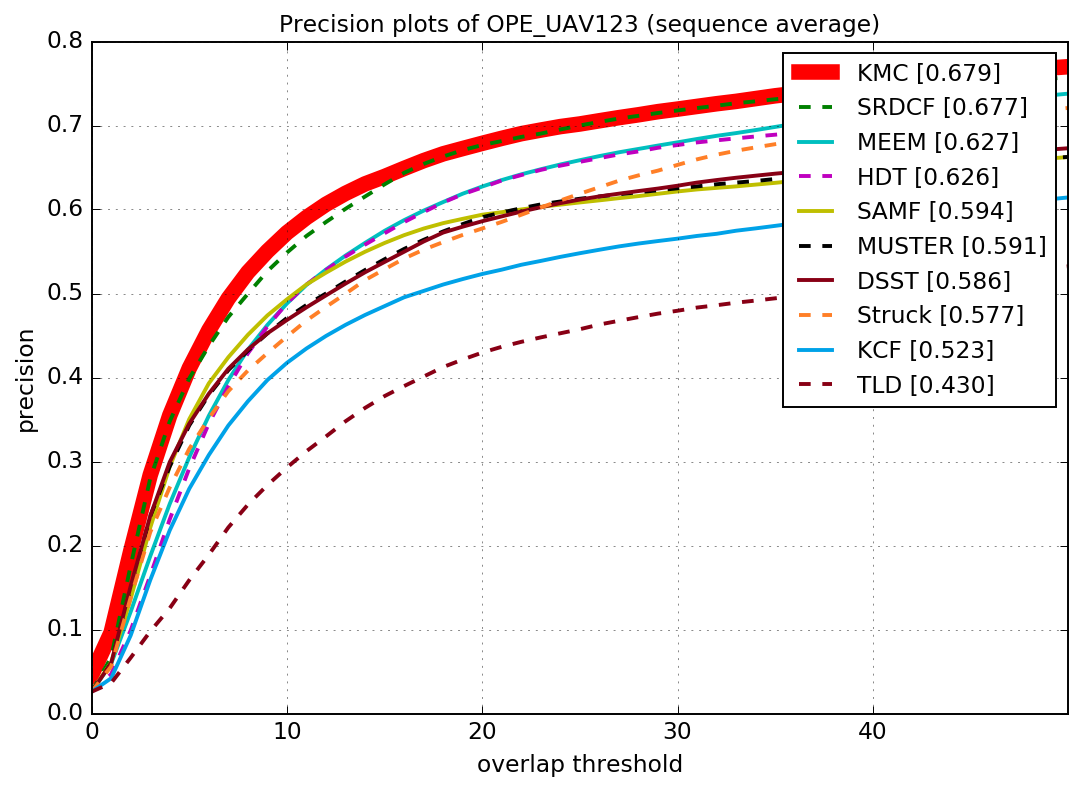}
        \caption{\small{Precision plot for UAV123 Dataset}}
        \end{subfigure}%
        \begin{subfigure}[c]{0.4\textwidth}
        \centering
            \includegraphics[width=7cm,height=5cm, clip]{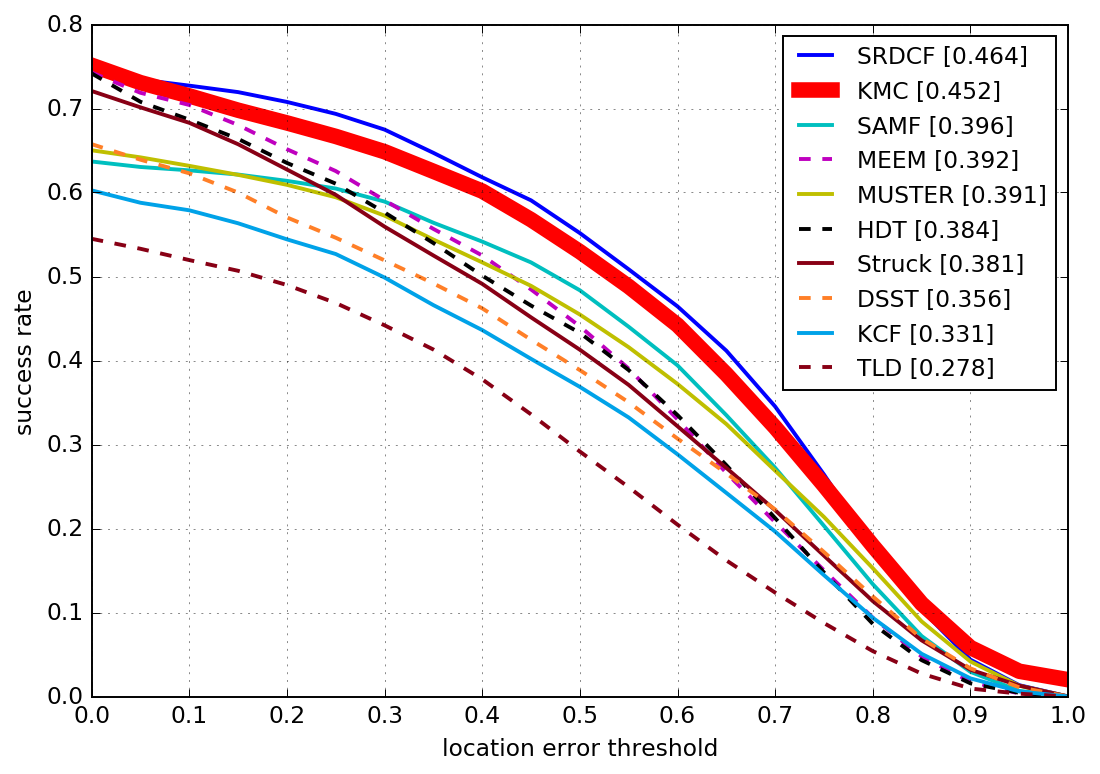}
            \caption{\small{Success plot for UAV123 Dataset}}
        \end{subfigure}

\caption{Precision plot and success plot of the state-of-the-art trackers for OTB-2015 and UAV123 dataset.
}
\label{fig:precision_plot_and_success_plot_OBT100}
\end{figure*}

We validate our KMC tracker in a comprehensive comparison with 10 state-of-the-art trackers:
HDT~\cite{qi2016hedged}, SRDCF~\cite{danelljan2015iccv}, MEEM~\cite{meem2014}, MUSTER~\cite{hong2015multi}, SAMF~\cite{samf}, DSST~\cite{danelljan2017dsst_tpami}, KCF~\cite{henriques2015high}, Struck~\cite{struck}, TLD~\cite{tld}, DCF~\cite{henriques2012exploiting}.

\vspace{\baselineskip}
\noindent \textbf{OTB-2015 Dataset:}
OTB-2015 dataset contains 100 video sequences and is the superset of OTB-2013~\cite{wu2013online} dataset which contains the original 50 video sequences.
The Results on the OBT dataset is shown in the top row of Fig.~\ref{fig:precision_plot_and_success_plot_OBT100}.
It can be seen that our tracker performs on par with a range of state-of-the-art trackers across two evaluation metrics. The most similar tracker is HDT where both correlation filter and deep features are adopted. However, HDT has heuristically chosen a set of higher level deep feature layers whereas in our approach, all layers before max pooling are universally chosen.
The SRDCF tracker is better than our tracker in the success metric due to heavy regularisation for negative training examples in SRDCF tracker. Hence it enables larger search area during tracking. Furthermore, because SRDCF tracker is able to search for larger area during detection ($4^2$ vs. our $3.2^2$ the target size), it indirectly manages the training set to handle the scenarios of occlusion and drift, of which the proposed KMC has been mostly penalised.

\vspace{\baselineskip}
\noindent \textbf{UAV123 Dataset:}
Visual tracking on unmanned aerial vehicles (UAVs) is a very promising application, since the camera can follow the target based on visual feedback and actively change its orientation and position to optimise the tracking performance. This UAV123 dataset contains a total of 123 video sequences and more than 110K frames.
The major difference between this UAV123 dataset and other popular tracking datasets is the effect of camera viewpoint change arising from UAV motion, the variation in bounding box size and aspect ratio with respect to the initial frame, and longer tracking sequences on average due to the availability of mounted camera moving with the target.

Results are shown in the bottom row of Fig.~\ref{fig:precision_plot_and_success_plot_OBT100}. It can be seen that due to heuristically chosen layers tuned to the OBT dataset in HDT, its performance suffers on this UAV123 dataset.  It worths noting that our trained multi-resolution convnet has never seen any image in this dataset. Moreover, no pesky threshold, hyperparameters nor weights are altered when applying the trained KMC network on the unseen UAV123 dataset, proving its ability to generalise across various tracking scenarios.

\section{Conclusions}

In this paper, we propose a novel kernelised multi-resolution convnet tracking algorithm that utilises the intermediate response maps from the kernelised correlation filter outputs.
The multi-resolution convnet learns the implicit translational output accurately and later an adaptive learning scheme is adopted for model update.
The learning paradigm is able to generalise across various datasets without change of hyperparameters. Moreover, it opens the door on the end-to-end temporal deep learning. Future works include better regularisation method as in~\cite{danelljan2015iccv} for negative instance mining, multi-layer fusion~\cite{TNNLS_wu2013} and incorporating recurrent nets on top to model temporal dynamics.

\section*{Details of the Code}
\noindent The python based code can be found at: \\
\footnotesize{\verb+https://github.com/stevenwudi/KMC_cvprw_2017+}

\section*{Acknowledgment}
\noindent This project has received funding from National Natural Science Foundation of China (NSFC) (61401287); Natural Science Foundation of Shenzhen (JCYJ20160307154003475, JCYJ2016050617265125).

{\small
\bibliographystyle{ieee}
\bibliography{egbib}
}
\end{document}